\documentclass[pdflatex,sn-mathphys-num]{sn-jnl}

\usepackage[utf8]{inputenc}
\usepackage{graphicx}%
\usepackage{multirow}%
\usepackage{amsmath,amssymb,amsfonts}%
\usepackage{amsthm}%
\usepackage{mathrsfs}%
\usepackage[title]{appendix}%
\usepackage{xcolor}%
\usepackage{textcomp}%
\usepackage{manyfoot}%
\usepackage{booktabs}%
\usepackage{algorithm}%
\usepackage{algorithmicx}%
\usepackage{algpseudocode}%
\usepackage{listings}%
\usepackage{enumitem}
\usepackage[utf8]{inputenc}
\usepackage{array}
\usepackage{tabularx}
\usepackage{longtable}
\usepackage{ragged2e} 
\usepackage{pdflscape}
\usepackage{lscape}
\usepackage{adjustbox}
\usepackage{float}
\usepackage{rotating}
\usepackage{makecell}
\usepackage[none]{hyphenat}

\theoremstyle{thmstyleone}%
\theoremstyle{thmstyletwo}%

\theoremstyle{thmstylethree}%

\begin{document}

\title[A Narrative Review of CDSS in Offloading Footwear for DFU] {A Narrative Review of Clinical Decision Support Systems in Offloading Footwear for Diabetes-Related Foot Ulcers}


\author*[1]{\fnm{Kunal} \sur{Kumar}}\email{kukumar@csu.edu.au}

\author*[1]{\fnm{Muhammad Ashad} \sur{Kabir}}\email{akabir@csu.edu.au}

\author[2]{\fnm{Luke} \sur{Donnan}}\email{ldonnan@csu.edu.au}

\author[3]{\fnm{Sayed} \sur{Ahmed}}\email{sayed@footbalancetech.com.au}

\affil[1]{\orgdiv{School of Computing, Mathematics and Engineering}, \orgname{Charles Sturt University}, \city{Bathurst}, \postcode{2795}, \state{NSW}, \country{Australia}}

\affil[2]{\orgdiv{School of Allied Health, Exercise and Sports Sciences}, \orgname{Charles Sturt University}, \city{Albury}, \postcode{2640}, \state{NSW}, \country{Australia}}

\affil[3]{\orgdiv{Foot Balance Technology Pty Ltd}, \city{Castle Hill}, \postcode{2154}, \state{NSW}, \country{Australia}}



\abstract{
\textbf{Background}  
Offloading footwear is a critical intervention for preventing and treating diabetic foot ulcers (DFUs), as it reduces plantar pressure (PP) and promotes healing. However, decision-making around its prescription remains fragmented, with variability in feature selection, limited personalization, and diverse evaluation approaches. While guidelines, knowledge-based systems, and machine learning (ML) applications have explored elements of decision-making, no clinical decision support system (CDSS) currently exists to provide feature-level prescriptions.  

\textbf{Methods}  
We conducted a narrative review of 45 studies, including twelve guidelines/protocols, 25 knowledge-based systems, and eight ML applications. Studies published from inception to August 2025 were included. Studies were thematically analyzed by type of knowledge, decision logic, evaluation methods, and associated technologies. 

\textbf{Results}  
Guidelines emphasize PP thresholds ($\leq$200 kPa or $\geq$25–30\% reduction) but lack detailed outputs. Knowledge-based systems apply rule-based and sensor-driven logic, integrating PP monitoring, adherence tracking, and usability testing. ML applications introduce predictive classification, optimization, and generative models, achieving high computational accuracy but with limited explainability and clinical validation. Evaluation practices remain fragmented: biomechanical testing dominates protocols, usability and adherence assessments are common in knowledge-based systems, while ML studies emphasize technical accuracy with limited linking to long-term clinical outcomes. From this synthesis, we propose a five-component framework for CDSS development that discusses: (1) minimum viable dataset, (2) hybrid decision architecture combining rules, optimization, and explainable ML, (3) structured feature-level outputs, (4) continuous validation and evaluation, and (5) integration into clinical and telehealth workflows.  

\textbf{Conclusions}  
The proposed framework presents a pathway to transform fragmented approaches into scalable, patient-centered CDSSs. Prioritizing interoperable datasets, explainable models, and outcome-driven evaluation will be critical for clinical adoption and improved DFU care.  
}




\maketitle

\section{Introduction}
\label{sec1}
Offloading footwear is a critical intervention for preventing and treating diabetes-related foot ulcers (DFUs), as it reduces plantar pressure (PP) and facilitates ulcer healing~\citep{RN9}. Evidence consistently demonstrates that footwear features such as rocker sole geometry, outsole density, and insole materials directly influence plantar pressure (PP) and ulcer recurrence risk~\citep{preece2017optimisation, ahmed2020footwear}. For example, moving the rocker apex distally and reducing the rocker angle has been shown to achieve clinically meaningful PP reductions~\citep{preece2017optimisation}. Similarly, custom-made insoles and rocker soles have demonstrated significant offloading effects, while intelligent insoles with real-time feedback reduce exposure to high-pressure activities and lower recurrence risk~\citep{abbott2019innovative, chatwin2021intelligent}. These findings underscore the importance of feature-level adjustments to footwear for effective DFU management.  

The clinical effectiveness of offloading footwear depends not only on biomechanical outcomes such as PP reduction but also on patient adherence. Suboptimal adherence, often ranging from 22\% to 60\% of daily wearing time, is insufficient to support healing of DFUs~\citep{racaru2022offloading}. In contrast, adherence above 80\% is associated with significantly reduced recurrence and faster healing~\citep{racaru2022offloading}. Barriers to adherence include patient preferences, financial limitations, and broader socioeconomic circumstances~\citep{cheema2020adherence, ababneh2024adherence}. Structured consultations have been proposed to improve adherence, yet they increase demands on healthcare providers (HPs) without conclusive evidence of improved clinical outcomes~\citep{racaru2022offloading}. This highlights the need for systems that can simultaneously personalize footwear prescription and reduce cognitive burden on HPs.  

Clinical decision support systems (CDSSs) offer a promising solution, having demonstrated value in other diabetes and DFU contexts. For example, CDSSs have improved glycemic control through expert-driven insulin dosing~\citep{contreras2018artificial}, enhanced DFU wound care risk assessment~\citep{araujo2020clinical}, and optimized hospital workflows to reduce complications and costs~\citep{garces2023clinical, ting2021health}. These examples show the potential of CDSSs to process complex, multidimensional data and deliver evidence-based, real-time recommendations.  

Similar reviews have explored technological innovations in offloading footwear. A recent narrative review mapped the use of sensors, smart insoles, and wearable systems for offloading but did not address how these technologies could be structured into decision-support frameworks for clinical use~\citep{bus2024offloading}. Another review highlighted advances in sensor-based monitoring for diabetic foot care, emphasizing pressure, temperature, and motion detection, but without exploring their integration into decision-making logic for footwear prescription~\citep{srass2023adherence}. These studies establish a strong technological foundation but leave open the critical question of how decision-making approaches from guidelines, knowledge-based systems, and machine learning (ML) can be adapted into CDSSs that output actionable, feature-level footwear prescriptions.  

Currently, no operational CDSS exists to support offloading footwear prescription by generating specific design features such as rocker angle or insole stiffness based on individual patient data. A preliminary search of major databases (PubMed, Scopus, Cochrane, and Web of Science) confirmed the absence of systematic or narrative reviews dedicated to this problem. While related reviews focus on glycemic management, risk prediction, or general DFU prevention, they do not address the specialized needs of footwear design~\citep{garces2023clinical, ting2021health}. Emerging prototypes and conceptual frameworks suggest potential~\citep{ahmed2023ai, bus2024offloading}, but these remain fragmented and unvalidated.  

This gap presents an opportunity for a targeted review that synthesizes existing decision-making approaches across guidelines, knowledge-based systems, and ML applications, critically appraises their decision logic and evaluation methods, and assesses their applicability to developing clinically viable CDSSs for offloading footwear prescription in DFU care.  

\subsection{Objective}
The objective of this study is to synthesize and critically appraise existing decision-making approaches, including guidelines, knowledge-based systems and machine learning applications, that generate feature-level prescriptions for offloading footwear or orthoses, and to evaluate their applicability for developing CDSSs in diabetic foot ulcer care.

\section{Methods}
\label{sec:methods}

\subsection{Study Design}
This study conducts a narrative review to explore computational decision-support methodologies that generate offloading footwear or orthosis prescription at the features level. The study does not assess methodical quality of the included studies as the emphasis of the review is on decision logic, system design, and feature outputs, rather than effect size of intervention efficacy. Additionally, since there are not many studies available in this domain, applying a strict quality filter may unintentionally exclude conceptual or early prototype that are highly relevant for understanding the research domain and informing future development.

\subsection{Information Sources and Search Strategy}

We searched PubMed, Embase, CINAHL, Cochrane Library, IEEE Xplore, ACM Digital Library, Scopus, and Web of Science from inception to August 2025. Grey literature sources included studies from arXiv, medRxiv, and conference proceedings.

The search strategy combined three conceptual blocks using the AND operator, with synonyms within each block linked using OR:

\begin{itemize}
    \item \textbf{Decision-support concepts:} "clinical decision support" OR "decision support system*" OR "expert system*" OR "rule-based" OR "algorithm*" OR "machine learning" OR "smart" OR "intelligent" OR "personali*"
    \item \textbf{Prescriptive footwear/orthosis terms:} ("therapeutic footwear" OR "offloading footwear" OR "custom shoe*" OR orthos* OR insole* OR "rocker sole*") AND (prescrib* OR recommend* OR design* OR optimis* OR "decision rule*")
    \item \textbf{Target population:} "diabetic foot" OR "foot ulcer" OR neuropath* OR "high risk foot"
\end{itemize}

Proximity operators were applied to identify transferable systems that generate actionable prescriptions. For example, the term rocker NEAR/3 insole was used to ensure that related concepts appeared in close proximity. Additionally, sentinel studies, including rocker optimization trials and intelligent insole prototypes, were captured using wildcard terms such as  "personali*" and optim* and prescrib*.

\subsection{Eligibility Criteria}
Studies were selected from inception to August 2025. Studies were eligible for inclusion if they involved adult populations with diabetes, peripheral neuropathy, a history of DFU, or biomechanically comparable risk profiles such as individuals with foot deformities or altered gait patterns. Healthy cohorts were also included when involved in proof-of-concept studies related to footwear prescription. The intervention had to include one or more computational methods such as rule-based, knowledge-based, optimization, or ML or artificial intelligence (AI) techniques, that generated a footwear or orthosis prescription or a defined feature set. Furthermore, studies that evaluated or facilitated the development or optimization of offloading footwear were also included. Studies had to report at least one outcome, including biomechanical measures (example, peak plantar pressure, pressure–time integral), clinical outcomes (example, ulcer incidence or recurrence), and system-level metrics (example, decision time, number of iterations), as well as usability, feasibility, or safety. Studies were excluded if they did not produce footwear outputs, lacked computational logic, or measurement-only devices that did not include a prescriptive module. Additionally, studies that solely employ engineering methodologies, such as Finite Element Modeling (FEM), were excluded, as these approaches fall outside the scope of applied computer science and do not involve decision-support logic.

\subsection{Study Selection and Data Extraction}
Study screening and data extraction was conducted by a team of researchers (KK, AK, LD). Potential studies were imported to Covidence for screening, where two reviewers (KK and AK) independently screened each article's title and abstract. Articles marked for inclusion by both reviewers were moved to full-text screening, which was subsequently completed by the two reviewers (KK and AK). Any conflicts during the screening process were resolved by the third reviewer (LD). After screening, KK extracted study characteristics using the following template:
\begin{itemize}
    \item Study type (guideline, protocol, knowledge-based system, ML application)
    \item Decision logic or recommendations 
    \item Evaluation techniques (validation of guidelines, protocols, systems, ML models, and evaluating footwear)
    \item Associated technologies (the type of hardware and software used to develop the system)
\end{itemize}

\subsection{Data Analysis}
The data analysis thematically organizes and discusses the findings based on the type of study and the underlying decision logic. The analysis focuses on identifying key components of CDSS development and the associated challenges, which are then used to structure a roadmap for future implementation in offloading footwear prescription for DFU care. Evidence from this synthesis further informs a comprehensive framework that integrates these components into defined CDSS layers, outlining the techniques and technologies currently available in the literature.
\section{Current Evidence} 
\begin{figure}[H]
\centering
\includegraphics[width=\linewidth]{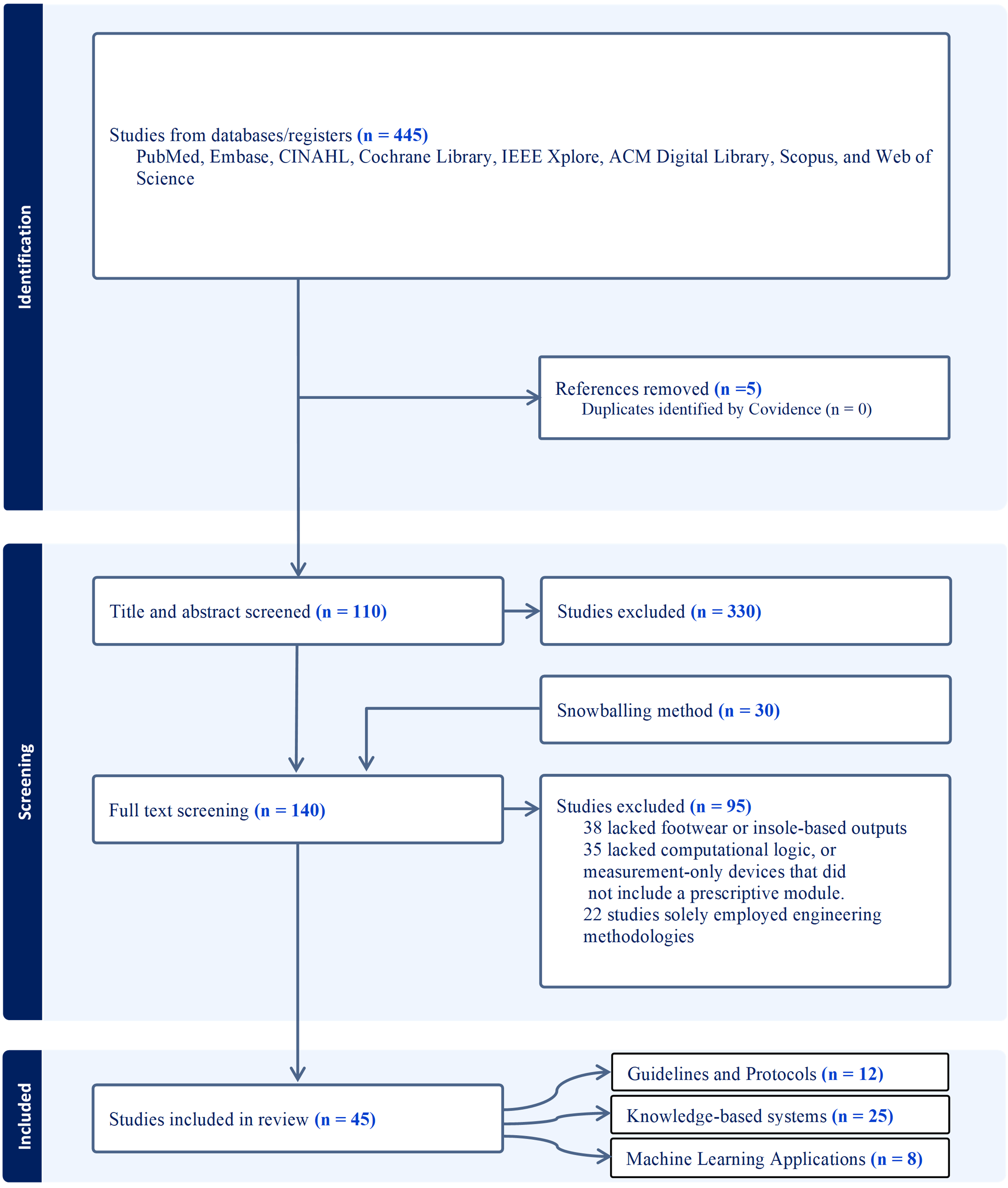}
\caption{PRISMA diagram}
\label{fig:prisma}
\end{figure}
A total of 45 studies were selected for final analysis, divided into three distinct groups as shown in Figure~\ref{fig:prisma}. In total, 95 studies were excluded from the review: 38 that did not focus on prescribing footwear or orthosis interventions, 35 that lacked sufficient computational or decision-making logic, and 22 that focused exclusively on engineering methodologies such as FE modeling. The studies were categorized into three major groups: (1) guidelines and protocols, (2) knowledge-based systems, and (3) machine learning (ML) applications, each offering unique approaches to assist in footwear prescription. 

\subsection{Guidelines and Protocols}
Guidelines and protocols provide structured information that can be useful in designing algorithms for decision making in CDSSs. Guidelines are documents that partially or fully provide recommendations for offloading footwear prescription, while protocols are structured, detailed procedures that delineate care in practice, such as a hospital protocol requiring inshoe PP testing before approving custom footwear. Table~\ref{tab:guideandprotocols_summary} summarizes these guidelines and protocols, outlining their decision logic or guidelines, techniques for validating the study and the footwear, and associated technologies for studies that applied the guidelines or protocols in a digital application.

{\RaggedRight
\begin{longtable}{p{1.0cm} p{2.0cm} p{3.0cm} p{3.0cm} p{2.0cm}}
\caption{Summary of guidelines and protocols used for decision-making in offloading footwear prescription for DFU patients}
\label{tab:guideandprotocols_summary} \\
\toprule
\textbf{Study} & \textbf{Type of Document} & \textbf{Decision Logic/ Guide} & \textbf{Validation Techniques} & \textbf{Associated Technologies} \\
\midrule
\endfirsthead

\toprule
\textbf{Study} & \textbf{Type of Document} & \textbf{Decision Logic/ Guide} & \textbf{Validation} & \textbf{Associated Technologies} \\
\midrule
\endhead

\bottomrule
\endlastfoot

~\citep{schaper2024practical} & Guideline (IWGDF 2023)
&
This guideline provides clear clinical advice on the appropriate use of offloading and therapeutic footwear.
&
\textbf{Guideline validation:} → Systematic literature review + expert consensus using GRADE methodology.

\textbf{Footwear validation:} → in-shoe PP measurements, with iterative modifications if needed.
&
N/A
\\
~\citep{ahmed2023ai} & Guideline
&
Assess risk, foot morphology and plantar biomechanics (plantar pressure), map findings to personalized footwear/insole features, then iteratively refine using in-shoe pressure testing and adherence monitoring until targets are met.
&
\textbf{Guideline validation:} → Systematic literature review + Retrospective clinical audit + Pedorthists’ + N-of-1 trials

\textbf{Footwear validation:} → in-shoe PP measurements, with iterative modifications if needed.
&
N/A
\\
~\citep{sousa2023three} & Protocol
&
Follow a three-step process: (1) screen the patient for risk factors, (2) provide appropriate offloading footwear or insoles, and (3) monitor and adjust prescriptions over time.
&
\textbf{Protocol validation:} → Clinical practice experience + literature evidence.

\textbf{Footwear validation:} → in-shoe PP measurements + patient followup, clinical validation by HP.
&
N/A
\\
~\citep{lopez2022clinical} & Protocol
&
Use a 3D foot scanner app to guide therapeutic footwear fitting, which reduces ill-fitting shoes and supports ulcer prevention in high-risk diabetic patients.
&
\textbf{Protocol validation:} → randomized controlled trial (RCT).

\textbf{Footwear validation:} → rate of ill-fitting footwear and ulcer recurrence.
&
→ Mobile app (3D scanner app) \newline
→ Smartphone camera (18–24 photos per foot). \newline
→ Database of therapeutic footwear models 
\\
~\citep{bus2020state} & Protocol
&
The protocol provides evidence- and consensus-based rules for designing custom-made footwear and insoles, and a 10-step pressure-relief algorithm to reduce plantar pressure in high-risk diabetic patients.
&
\textbf{Protocol validation:} → Systematic literature review + expert consensus

\textbf{Footwear validation:} → in-shoe PP measurements, with follow-up assessments.
&
N/A
\\
~\citep{zwaferink2020optimizing} & Protocol
&
Custom-made footwear should be designed using plantar pressure measurements, 3D foot shape, and scientific-based modifications to ensure forefoot peak pressures remain below 200 kPa for ulcer prevention.
&
\textbf{Protocol validation:} → Footwear comparison

\textbf{Footwear validation:} → in-shoe PP measurements
&
N/A
\\
~\citep{bus2020guidelines} & Guideline (IWGDF 2019)
&
This guideline provides clear clinical advice on the appropriate use of offloading and therapeutic footwear.
&
\textbf{Guideline validation:} → Systematic literature review (meta-analysis) + expert consensus using GRADE methodology.

\textbf{Footwear validation:} → N/A
&
N/A
\\
~\citep{van2018diabetic} & Guideline (Australian Guideline)
&
People with diabetes should wear appropriately fitting, protective, offloading footwear (e.g., depth/width shoes, custom insoles) to reduce plantar pressure, prevent ulceration, and improve adherence; high-risk patients should be referred for custom-made footwear.
&
\textbf{Guideline validation:} → literature review + expert consensus.

\textbf{Footwear validation:} → N/A
&
N/A
\\
~\citep{australia2016australian} & Guideline
&
It provides a standardized dataset to guide DFU assessment, management and outcome monitoring so services can benchmark care and improve quality.
&
\textbf{Guideline validation:} → literature review (guidelines) + DFU datasets + expert consensus.

\textbf{Footwear validation:} → N/A
&
N/A
\\
~\citep{giacomozzi2013learning} & Protocol
&
The protocol tests diabetic footwear using Pedar insole pressure and gait line measurements, comparing results with predefined thresholds to decide if footwear is safe, needs modification, or should be rejected. 
&
\textbf{Protocol validation:} → literature review (guidelines) + DFU datasets + expert experience, and reference data collected from healthy volunteers

\textbf{Footwear validation:} → in-shoe PP measurements
&
N/A
\\
~\citep{bus2011evaluation} & Protocol
&
Therapeutic footwear is iteratively modified (e.g., insole adjustments, pads, rocker sole changes) and tested with in-shoe plantar pressure analysis until high-risk regions achieve $\geq$25\% pressure reduction or pressures $<$200 kPa .
&
\textbf{Protocol validation:} → clinical trial

\textbf{Footwear validation:} → in-shoe PP measurements
&
N/A
\\
~\citep{dahmen2001therapeutic} & Protocol
&
Match foot pathology (sensory loss, limited joint mobility, deformities, Charcot foot, amputations, ulceration) with specific footwear adaptations (insole type, shoe height, outsole pivot point, rigidity, heel/shock absorption).
&
\textbf{Protocol validation:} → literature, expert opinion, and clinical experience

\textbf{Footwear validation:} → N/A
&
N/A
\\
\end{longtable}
}

A total of 12 documents were selected for review in this section, comprising five guidelines and seven protocols. Two of the guidelines are international, namely the IWGDF 2019~\citep{bus2020guidelines} and its updated version, the IWGDF 2023~\citep{schaper2024practical}. While both guidelines are based on systematic literature reviews and expert consensus using the GRADE methodology, the 2023 update additionally recommends validating prescribed footwear using PP thresholds. Two other guidelines originate from Diabetic Foot Australia (DFA): one outlines a minimum dataset for the assessment, management, and outcome monitoring of DFU~\citep{australia2016australian}, while the national Australian guideline~\citep{van2018diabetic} provides recommendations on appropriate and protective offloading footwear. Another study presents a guideline for data to consider for prescribing offloading footwear with a specific focus on personlized footwear development and improving adherence~\citep{ahmed2023ai}. All studies were validated through literature reviews and expert consensus, while some additionally used DFU datasets, clinical trials, and HPs experience for validation. 

The seven protocols offer detailed steps for assessing, prescribing, and evaluating offloading footwear. Two protocols emphasize evaluating patient conditions—such as risk factors, sensory loss, joint mobility, deformities, and comorbidities—to prescribe appropriate therapeutic footwear~\citep{dahmen2001therapeutic, sousa2023three}. The remaining five studies develop offloading footwear using PP-guided techniques, whereby footwear is iteratively modified to achieve a PP $\leq$ 200 kPa or a $>$30\% pressure reduction in high-risk regions~\citep{lopez2022clinical, bus2020state, zwaferink2020optimizing, giacomozzi2013learning, bus2011evaluation}. Notably, only one study implemented the protocol in a mobile application, using 3D foot scans to improve the fit of prescribed footwear~\citep{lopez2022clinical}. 
Evaluation methods across the studies primarily relied on systematic literature reviews and expert consensus, with some incorporating clinical trials and DFU datasets.

PP measurement was the predominant method used for evaluating the effectiveness of offloading footwear across the reviewed studies. Several protocols and guidelines employed in-shoe PP analysis to guide iterative footwear modifications and validate pressure reduction outcomes. In addition to PP-based assessment, two studies incorporated alternative evaluation methods such as ill-fitting footwear rates, clinical validation by health professionals, and ulcer recurrence tracking as supplementary indicators of therapeutic efficacy. Knowledge derived from these guidelines and protocols will serve as foundational data sources for designing decision logic and establishing evaluation criteria in the development of CDSSs for offloading footwear prescription.

\subsection{Knowledge-based Systems}
While the previous section focused mainly on studies that did not offer digital solutions to execute the decision logic, this section will focus on studies that apply the decision logic as part of a digital application. A total of 25 studies were selected in this category, making up the largest group of studies in the group as illustrated in Figure~\ref{fig:prisma}. The knowledge-based studies are presented in Table~\ref{tab:knowledge_study_summary}, outlining the type of knowledge, the source of knowledge, decision logic, decision evaluation criteria, and associated technologies used in the implemented systems. 

{\RaggedRight
\begin{longtable}{p{1.0cm} p{1.8cm} p{2.8cm} p{2.8cm} p{2.8cm}}
\caption{Summary of Knowledge-based studies implementing decisions in offloading footwear prescription for DFU patients}
\label{tab:knowledge_study_summary} \\
\toprule
\textbf{Study} & \textbf{Knowledge Type and Source} & \textbf{Decision Logic} & \textbf{Decision Evaluation} & \textbf{Associated Technologies} \\
\midrule
\endfirsthead

\toprule
\textbf{Study} & \textbf{Knowledge Type and Source} & \textbf{Decision Logic} & \textbf{Decision Evaluation} & \textbf{Associated Technologies} \\
\midrule
\endhead

\bottomrule
\endlastfoot

~\citep{cay2025smartboot} & Rule-based \newline
source: sensors and literature &
Footwear sensors objectively measure adherence based on predefined thresholds and send notifications immediately to patients' smartwatch and the HPs' cloud dashboard. &
→ Adherence detection accuracy: 96–97\% \newline
→ TAM survey: high perceived usefulness, ease of use, and intent to adopt &
→ Smart Footwear (Inertial Measurement Unit sensor) \newline
→ Smartwatch \newline
→ Cloud computing technologies for online dashboards for HPs \newline
→ Validation with LEGSys gait analyzer and observer logs \newline
→ Usability assessment with TAM survey \\

~\citep{van2025short} & Behavioral model \newline
source: sensors and literature &
Patient is systematically given structured education, motivational interviewing, and custom indoor footwear. The study uses the Groningen algorithm for performing wear-time analysis &
→ Adherence measured by wear time \newline
- mean wear time increased from 7.5 to 8.3 h/day \newline
- Low-adherence group improved (4.0 to 5.5 h/day) \newline
- Indoor footwear had strongest effect (+2.0 to +2.7 h/day, significant) &
→ Smart Footwear (inshoe temperature sensors) \\

~\citep{moulaei2025usability} & Rule-based \newline
source: sensors and literature &
Patients’ footwear is evaluated and notifications are sent to users’ smartphone based on thresholds set by clinicians &
→ Provided continuous monitoring of plantar pressure, temperature, and humidity &
→ Smart Footwear (inshoe pressure, temperature, and humidity sensors) \newline
→ Hardware (microcontroller – single-chip computer system, Bluetooth module, and batteries) \newline 
→ Android app (Android Studio 3.6.2, Bluetooth v2.0+EDR) \\

~\citep{ghazi2024design} & Rule-based \newline
source: sensors and literature &
Sensors check against set limits for gait and PP data and send alerts through the Internet of Things (IoT) network to HPs or patients &
→ Remote monitoring of PP and gait \newline
→ Patient adherence monitoring via mobile app \newline
→ HPs get continuous access to biomechanical data &
→ Smart Footwear (inshoe pressure, motion sensors) \newline
→ Hardware (IoT microcontroller and wireless communication module) \newline
→ Mobile application monitoring adherence \\

~\citep{havey2024adherence} & Rule-based \newline
source: sensors and literature derived thresholds &
The algorithms analyze signals from temperature, proximity, and motion sensors to detect patterns that indicate whether the boot is being worn or not, and the results are validated against wear-time diaries. &
→ Matthews Correlation Coefficient (MCC) used to compare algorithm performance with subject diaries \newline 
- Multi-sensor algorithms (i.e., combining proximity + temperature + acceleration) achieved high accuracy (MCC = 0.96) compared to diaries \newline 
- Outperformed single-sensor approaches &
→ Smart Footwear (inshoe pressure, motion sensors) \newline
→ Hardware (microcontroller, real-time clock, memory (22 days storage), battery, sensor suite) \newline
→ Data: time-stamped per minute, multiple sensor streams \\
~\citep{malki2024plantar} & Rule-based \newline
source: sensors and literature &
Predefined rule-based geometric parameters (rocker radius, apex position, heel cup geometry) guide adjustments to the self-adjusting insole to achieve optimal pressure reduction &
→ PP evaluated for distal heel, forefoot, and toe &
→ Smart Footwear (inshoe pressure sensors) \newline
→ Matlab + SolidWorks + 3D printing (used to design individualized rocker midsoles using rule-based algorithmic parameters) \newline
→ Self-adjusting insole with hexagonal buckling forefoot elements. 
→ Gait lab treadmill setup. \\
~\citep{vossen2023integrated} & Rule-based \newline
source: sensors and literature &
Footwear is adjusted until plantar pressure is reduced to $\leq$200 kPa or by at least 25\%, while temperature monitoring triggers alerts if a hotspot ($\leq$2.2 °C) persists for two days, leading to advice on reducing activity and referral if it continues. &
→ Planned to assess PP \newline
→ Questionnaires concerning
quality of life, costs, disease, and self-care knowledge; physical activity and footwear use monitoring \newline
→ Clinical
monitoring for foot ulcer outcomes.
&
→ Smart Footwear (inshoe pressure, and temperature sensors) \newline
→ Wearable activity monitor
\\
~\citep{jarl2023personalized} & Rule-based \newline
source: sensors, literature, RCTs and guidelines &
Personalized offloading footwear designs created by evaluating PP reduction, weight-bearing control, and adherence monitoring with inshoe sensor data  &
→ Evaluate PP \newline
→ Monitor adherence \newline
→ Evaluate footwear design (advocates lighter, more acceptable, and adaptive footwear) \newline
→ Real-time monitoring of patient data.
&
→ Smart Footwear (inshoe pressure, and temperature sensors) \newline
→ Hardware (3D scanning, CAD, 3D printing)
\newline
→ Wearable activity monitor \newline
→ Clinical integration of personalized medicine principles.
\\
~\citep{hemler2023intelligent} & Rule-based \newline
source: sensors and literature &
Pressure data from in-shoe sensors are compared to thresholds, and fluid modules are activated or deactivated to offload high-pressure areas, with future systems aiming for adaptive ML-based control. \newline
&
→ Evaluate PP (Demonstrated 18–32\% PP reduction in pilot walking tests.) \newline
→  Feedback from patients, caregivers, and clinicians (95\% found use beneficial).
&
→ Smart Footwear (inshoe pressure sensors) \newline
→ Hardware (insole housing with pressure sensors, MR fluid modules, batteries, microcontroller)
\newline
→ Algorithms: pressure-threshold rules \newline
→ Usability: daily charging (~9000 mAh battery capacity), footwear aesthetics designed for patient acceptability. 
\\
~\citep{malki2024effects} & Rule-based \newline
source: sensors and literature &
An optimization algorithm adjusts rocker midsole parameters and self-adjusting insole parameters to achieve biomechanical targets for pressure offloading. \newline
&
→ PP reduction (24–48\%) in hallux, toes, central/lateral forefoot.
&
→ Smart Footwear (inshoe pressure sensors) \newline
→ Foot/ankle axis and deformity assessment devices \newline
→ Shoe dimensions and dermal assessment tools \newline
→ Matlab-based algorithm to define rocker parameters. \newline
→ Hardware (3D-printing of midsoles and hexagonal TPU insoles)
\\
~\citep{mancuso20233d} & Rule-based \newline
source: sensors, literature and Multidisci-\newline plinary input (clinician, podiatrist, engineer, and patient)
&
Plantar pressure hot spots are identified and linked to footwear design modifications, which are iteratively adjusted until pressures fall below 200 kPa or are reduced by more than 30\%. \newline
&
→ PP reduction by $>$50\% at hindfoot \newline
→ Timed Up and Go (TUG) performance measured patients mobility and independence (improved by 12.5s vs 14.1s for barefoot) \newline 
→ Patient feedback on comfort and balance. 
&
→ Smart Footwear (inshoe pressure sensors) \newline
→ Mobile app for patient scanning and data collection. \newline
→ Hardware (CAD software, 3D printer and EVA antibacterial sheet covering) 
\\
~\citep{tang2023wearable} & Rule-based \newline
source: sensors and literature &
Insoles and footwear are repeatedly adjusted and tested against plantar pressure targets, with modifications continued until pressures are reduced by more than 25\% or fall below 200 kPa. \newline
&
→ PP reduction (insole did not elevate PP) \newline
→ Shear reduction (reduced up to 82\%) \newline
→ Validation with multiple footwear types (trainers, plimsolls, therapeutic rocker-sole) 
&
→ Smart Footwear (inshoe TRIPS sensors + flexible cable to electronics hub) \newline
→ Hardware (Microcontroller + capacitance-to-digital converters, Onboard storage + wireless transfer, USB-C power, smartphone compatibility) 
\\
~\citep{panahi2022development} & Rule-based \newline
source: sensors and literature &
Insoles and footwear are repeatedly modified and tested against plantar pressure thresholds, with adjustments continued until they achieve either a $>$25\% pressure reduction or values below 200 kPa. \newline
&
→ PP ($>$25\% reduction at high-risk RoIs). 
&
→ Smart Footwear (inshoe pressure calibrated sensors with high sampling frequency)
\\
~\citep{park2023smart} & Rule-based \newline
source: sensors and literature &
If no movement is detected for 10 minutes or boot and watch step counts don’t match, the system classifies the boot as removed and sends a real-time alert to the patient’s smartwatch and clinician dashboard.
&
→ Adherence monitoring (acheived sensitivity of 90.6\%, Specificity of 88.0\%, and Accuracy of 89.3\% \newline
→ Balance monitoring (improved balance (reduced COM sway, large effect sizes) \newline
→ User feedback: device usability, comfort, nonintrusive, and innovativeness. 
&
→ Smart Footwear (inshoe pressure sensors) \newline
→ Smartwatch app with 4G LTE SIM for alerts + feedback \newline
→ Cloud-based clinician dashboard \newline
→ Butterworth filters (signal processing filter) \newline 
→ Algorithms for movement classification \newline
→ Wearable validation system for COM sway and gait \newline
→ Patient training for use and adherence
\\
~\citep{mbue2022benefits} & rule-based \newline
source: sensors and literature &
If plantar pressure at a sensor exceeds 35–50 mmHg for more than 15 minutes, an alert is sent to the patient’s smartwatch.
&
→ PP monitoring (Smart insole alone improved real-time offloading)
→ Adherence monitoring (Insole+education group showed highest adherence) 
&
→ Smart Footwear (inshoe pressure, and temperature +  RFID sensors)
→ Smartwatch with multimodal alerts. 
→ Structured education package (90-min session, pamphlet, tools, follow-up calls). 
\\
~\citep{beach2021monitoring} & Comparison- \newline based \newline
source: sensors and literature &
The system compares time constants in people with diabetes against controls to identify deviations that may serve as early biomarkers of diabetic foot ulcer risk.
&
→ Temperature monitoring (faster temperature rise times at hallux and 5th metatarsal head in diabetic group vs controls) 
&
→ Smart Footwear (temperature sensors) \newline
→ Hardware (Custom circuitry, microcontroller, 3D-printing of flexible TPU insoles with embedded sensors) \newline
→ Smartphone \newline
→ Python-based data analysis (fitting exponential and linear models).
\\
~\citep{chatwin2021intelligent} & Rule-based \newline
source: sensors, literature, and podiatry review &
The smart insole system uses predefined rule-based categories to monitor plantar pressures and deliver personalized feedback to the patient via a digital watch.
&
→ PP monitoring (reduced bouts of high plantar pressure after 16 weeks) 
&
→ Smart Footwear (pressure sensors) \newline
→ Digital display watch for audio-visual-vibrational alerts. \newline
→ Custom MATLAB scripts for data processing. \newline
→ Monthly calibration of sensors.
\\
~\citep{najafi2018cost} & Rule-based \newline
source: sensors and literature &
When plantar pressure rises above a predefined safety threshold, the insole triggers an alert that is delivered to the patient’s smartwatch through vibration and visual cues.
&
→ Ulcer recurrence rates (0.14 vs 0.62 at 18 months)
→ Cost Analysis (standard of care (SOC) + smart device saved US\$6,702 per ulcer avoided compared to SOC alone, Short-term cost saving around \$978 per event)
&
→ Smart Footwear (pressure sensors) \newline
→ smartwatch 
→ Decision tree (Manual not ML model) modeling tools
\\
~\citep{lutjeboer2018validity} & Rule-based \newline
source: sensors and literature &
Wear-time was classified using three algorithms: the Groningen method (slope/peak-based without a fixed cut-off), Algorithm-25 (temperature threshold of 25 °C), and Algorithm-29 (temperature threshold of 29 °C).
&
→ Adherence Monitoring (Validated using Groningen algorithm) 

&
→ Smart Footwear (inshoe temperature sensors) \newline
→ Hardware (100-day storage within sensor, RFID data transfer) \newline
→ Matlab scripts for Groningen algorithm \newline
→ Sports camera 
\\
~\citep{bencheikh2018low} & Rule-based \newline
source: sensors and literature &
The smart insole used rule-based decision logic, where real-time pressure, temperature, and humidity readings were compared against predefined thresholds, and any exceedance triggered immediate alerts to the patient via a smartphone app.
&
→ PP monitoring \newline
→ Temperature monitoring \newline
→ Humidity monitoring \newline
→ Cost analysis (low cost prototype) \newline
→ Real-time data monitoring by HPs
&
→ Smart Footwear (inshoe pressure, temerature, and humidity sensors) \newline
→ Hardware (microcontroller, multiplexer, amplifier, filters, batteries with regulators) \newline
→ Bluetooth module + smartphone application for visualization/alerts. 
\\
~\citep{najafi2017smarter} & Rule-based \newline
source: sensors and literature &
If plantar pressure remains above 35–50 mmHg for more than 15 minutes, an alert is triggered indicating the need to offload.
&
→ Adherence monitoring ($\leq$ 1 alert every 2 hours improves adherence) \newline
→ PP monitoring \newline
→ New ulcers (No new ulcers formed during study period )
→ User feedback (Users found device acceptable and useful)
&
→ Smart Footwear (inshoe pressure sensors) \newline
→ Smartwatch for alerts (audio, visual, vibration) \newline
→ MATLAB algorithm for processing sensor data \newline
→ Statistical analysis (ANOVA, Cohen’s d, correlations) \newline
→ Patient training on use and adherence monitoring
\\
~\citep{benbakhti2014instrumented} & Rule-based \newline
source: sensors and literature & 
The system triggers alerts whenever plantar pressure, local temperature, or humidity rise above predefined thresholds.
&
→ Monitoring real-time data and alerts
&
→ Smart Footwear (inshoe pressure, temperature, and humidity sensors) \newline
→ Hardware (Arduino Mini Pro, OpAmps, multiplexer, Bluetooth module, batteries with regulators \newline
→ Custom Android app
\\
~\citep{bernabeu2013cad} & Rule-based and machine learning \newline
source: predefined knowledge- \newline based rules, images of foot and show last &
The system integrates patient foot measurements, predefined design rules, and neural network predictions to generate customized diabetic shoe lasts, trained on datasets of diabetic and non-diabetic foot shapes.
&
→ Footwear fitting
&
→ Database of 150+ foot/last measurements \newline
→ Neural network models to infer relations not explicitly encoded \newline
→ Hardware (3D printing)
\\
~\citep{davia2011shoes} & Rule-based \newline
source: predefined rules + test cases & 
Knowledge-based rules link patient biomechanical variables and clinical data to specific footwear design features, such as rocker angle, heel height, and apex angle.
&
→ Footwear design time (Faster, rule-driven design with 64\% time savings compared to traditional process) 
&
→ Hardware (3D foot scanner + baropodometric platform + gait analysis, foot pressure maps and 3D foot mesh tools) \newline
→ Knowledge based repository (rules + test cases). \newline
→ XML-based data integration.
\\
~\citep{dabiri2008electronic} & Rule-based \newline
source: sensors, literature, rules & 
When plantar pressure rises above predefined safe limits, the system issues an alert that prompts the patient or clinician to take corrective action.
&
→ PP monitoring
&
→ Smart Footwear (inshoe pressure) \newline
→ Hardware (Wireless transmission hardware/software, External monitoring unit for data collection/alerts. 
→ Algorithm for threshold calibration.
\\
\end{longtable}
}

Across the 25 knowledge-based studies selected for review, rule-based decision-making emerged as the predominant approach, applied in 23 studies \citep{cay2025smartboot, moulaei2025usability, ghazi2024design, havey2024adherence, malki2024plantar, vossen2023integrated, jarl2023personalized, hemler2023intelligent, malki2024effects, mancuso20233d, tang2023wearable, panahi2022development, park2023smart, mbue2022benefits, chatwin2021intelligent, najafi2018cost, lutjeboer2018validity, bencheikh2018low, najafi2017smarter, benbakhti2014instrumented, davia2011shoes, dabiri2008electronic}. One study adopted a behavioral model \citep{van2025short}, while another combined rule-based knowledge with machine learning techniques \citep{bernabeu2013cad}. Given that most studies developed smart applications equipped with embedded sensors, the dominant sources of knowledge were sensor data and literature review (n = 20). A smaller number of studies integrated additional evidence sources, including RCTs and guidelines \citep{jarl2023personalized}, multidisciplinary team input \citep{mancuso20233d}, podiatry review \citep{chatwin2021intelligent}, or structured rules and datasets \citep{bernabeu2013cad, davia2011shoes, dabiri2008electronic}. 

In terms of decision logic, the majority of studies (n = 20) relied on predefined thresholds to trigger alerts or footwear modifications when plantar pressure, temperature, humidity, or adherence values exceeded safe limits. A smaller subset employed design optimization frameworks \citep{bernabeu2013cad, davia2011shoes, mancuso20233d}, comparative analytical methods such as temperature time-constant analysis \citep{beach2021monitoring}, or behavioral interventions centered on structured education and motivational interviewing \citep{van2025short}. 

Evaluation criteria most frequently focused on PP, reported in 16 studies, where footwear was iteratively modified to achieve thresholds of $\leq$200 kPa or reductions of $\geq$25–30\%. Adherence monitoring was the second most common criterion (n = 9), typically measured through wear-time sensors, validated against diaries, or through automated alerts provided to patients or healthcare professionals. Other evaluation methods included temperature monitoring (n = 6), humidity monitoring (n = 3), ulcer recurrence or prevention outcomes (n = 2), usability and patient feedback (n = 5), and cost-effectiveness analysis (n = 2). Additionally, unique measures such as patient mobility \citep{mancuso20233d}, footwear fitting quality \citep{bernabeu2013cad}, and design process efficiency \citep{davia2011shoes} were reported. Overall, the evidence demonstrates that current CDSS-related approaches in offloading footwear prescription are overwhelmingly rule-based and sensor-driven, with plantar pressure monitoring serving as the principal evaluation criterion across studies.

The findings from these studies provide a strong foundation for developing CDSSs for offloading footwear prescription. The dominance of rule-based approaches and sensor-derived data indicates that established thresholds, such as plantar pressure limits and adherence metrics, can be directly translated into CDSS decision logic. At the same time, the integration of evidence from guidelines, RCTs, and multidisciplinary input highlights opportunities to strengthen these systems with broader knowledge sources. Finally, the reported evaluation criteria—particularly plantar pressure outcomes, adherence, and usability—offer measurable benchmarks for validating CDSS performance. Together, these insights support the design of CDSSs that combine real-time sensor inputs, evidence-based rules, and patient-centered feedback to enable more personalized and clinically robust footwear prescriptions.

\subsection{Machine Learning Applications}
ML applications provide predictive capabilities that can enhance decision-making processes in healthcare technologies. This section examines the application of ML in offloading footwear prescription, with particular focus on the types of models employed, the underlying decision logic, the evaluation criteria, and the associated technologies. This information is presented in Table~\ref{tab:ML_summary}. A total of eight studies applied ML techniques within this domain for patients with DFU.

{\RaggedRight
\begin{longtable}{p{1.0cm} p{2.5cm} p{2.5cm} p{2.5cm} p{2.5cm}}
\caption{Summary of Machine Learning applications used for decision-making in offloading footwear prescription for DFU patients}
\label{tab:ML_summary} \\
\toprule
\textbf{Study} & \textbf{Model and Dataset} & \textbf{Decision Logic} & \textbf{Decision/ Model Evaluation} & \textbf{Associated Technologies} \\
\midrule
\endfirsthead

\toprule
\textbf{Study} & \textbf{Model and Dataset} & \textbf{Decision Logic} & \textbf{Decision/ Model Evaluation} & \textbf{Associated Technologies} \\
\midrule
\endhead

\bottomrule
\endlastfoot

~\citep{rajagopal2025personalized} & Model: CNN-GRU, Autoencoder + Random Forest, LSTM-\newline XGBoost Dataset: (1) Grayscale pressure sensor heat maps for foot posture (2) Clinically Validated Foot
Condition Dataset
and (3) Footwear Recommendation Dataset for Specific Foot Conditions \newline
&
AI generates multiple design concepts, and human designers choose those that best match medical requirements and patient preferences. &
→ Model Accuracy (Models achieved high accuracy: gait (98.3\%), disease (92.8\%), footwear recommendation (98.1\%)) \newline
→ Evaluate synthetic Footwear Images \newline
→ Patient and HP feedback (positive feedback on personalization and feasibility) &
→ ML models \newline
→ Generative AI tool \newline
→ Clinical validation by podiatrists/ \newline physiotherapists. 
\\
~\citep{resch2025improving} & Model:  Generative AI (MidJourney, Stable Diffusion) \newline Dataset: Images collected from standardized commercial orthoses devices, development concepts, research prototypes, and 3D-printed models.
 &
The pipeline first classifies gait using a CNN–GRU model, then identifies disease using an autoencoder with random forest, and finally recommends footwear features through an LSTM combined with XGBoost. Generative AI tool ( Stable Diffusion XL) to generate footwear images based on patient conditions &
→ Model Accuracy (Models achieved high accuracy: gait (98.3\%), disease (92.8\%), footwear recommendation (98.1\%))
→ Human evaluation of generated footwear 
→ HPs in the loop to ensure biomechanical safety of footwear
&
→ Generative AI models (trained on large-scale image datasets). 
→ Human designers for curation and evaluation.
\\
~\citep{lu2025parametric} & Model:  CNN + FCNN and Bayesian optimization \newline Dataset: simulation data generated using finite element (FE) method 
 &
The CNN extracts features from simulation data, the FCNN predicts insole performance, and Bayesian optimization selects the design that best reduces plantar pressure. &
→ PP evaluation (achieved up to 44.45\% reduction in plantar pressure and showed improved pressure distribution) \newline
→ Evaluate arch conformity \newline
→ Design time (Faster design cycle without repeated gait lab testing)
&
→ Generative AI models (trained on large-scale image datasets). \newline
→ FE software \newline
→ Computer aided design (CAD) tools \newline
→ ML frameworks \newline
→ Hardware (TPU, 3D printing) \newline
→ Experime- \newline ntal data (CT scans, gait force plates, fluoroscopy, PEDAR pressure insoles)
\\
~\citep{behuranew} & Model: Backpropagation Neural Network (BPNN) \newline Dataset: foot features (arch index, heel angle, leg angle) collected from patients 
 &
A BPNN predicts arch height and wedge parameters, which are exported as an STL file and used to CNC-mill an EVA insole. &
→ Model Evaluatrion (ANN achieved $R^2$ of 0.99, Mean Absolute Error (MAE) of 0.18–0.30. 
→ Design time (reduced design time)
→ PP evaluated (clinically validated with pressure reduction) 
&
→ ML models (BPNN (3-17-3 architecture) trained with 70/15/15 split). \newline
→ Hardware (3D foot scanners + anthropometric data, CNC milling machine for EVA insoles, Pressure mats for validation)
\\
~\citep{zhang2024prediction} & Model: Multilayer Perceptron (MLP) \newline Dataset: Collected barefoot ink footprints + dynamic in-shoe pressure 
 &
The MLP model uses low-cost ink footprint data to predict plantar pressure maps, which then guide material choices for insole design. 
&
→ Model Evaluation (MAE 3.57–5.51\% across insoles, ROC–AUC up to 95.1\% with localization embedding)   
→ PP evaluation (quantitative maps linked to specific insole materials)
&
→ ML models (Patch-based 7-layer MLP, ReLU, Adam optimizer, 400 epochs) \newline
→ Hardware (Ink footprint collection system, pressure sensors systems)\newline  
→ Dataset: 520 footprint-pressure pairs augmented to ~10,400 samples 
\\
~\citep{o2024ai} & Model: Random Forest \newline Dataset: collected plantar pressure data while performing five occupational tasks (standing, walking, pick and place, assembly, manual handling)  
 &
A Random Forest model classified different PP metrics into specific occupational tasks, while SHAP identified which features (peak pressure, Center of Pressure (CoP variability) most influenced the classification, ensuring interpretability.
&
→ Model Evaluation (Accuracy of 80 - 86\% \newline
→ Feature evaluation using SHAP visualizations
&
→ ML models (Random Forest classifier, SHAP library for explainability) \newline
→ Hardware (pressure sensors) \newline
→ Feature extraction pipeline (statistical, CoP, morphology). 
\\
~\citep{jung2022decision} & Model: Decision Tree (CART) \newline Dataset: public dataset containing 1548 patients diagnosed with pes planus 
 &
CART model used patient clinical and biomechanical variables to choose between gait plate or arch-support–heel cup orthoses.
&
→ Model Evaluation (Accuracy of 80.16\% \newline
→ Evaluate decision rules (generated 15 decision rules) \newline
→ Evaluate influential predictors (identified angle, age, subtalar eversion, and hip internal rotation)
&
→ ML models (Random Forest classifier, SHAP library for explainability) \newline
→ Large annotated dataset (1548 patients, reduced to 418 after cleaning) \newline
→ Clinical measurements devices \newline
→ Computing infrastructure
\\
~\citep{jones2019biometric} & Model: Artificial Neural Network (ANN) \newline Dataset: Not available (conceptual model only)
 &
ANN analyzes sensor data from pressure, shear, temperature, humidity, and movement to predict ulcer risk and send alerts when thresholds are exceeded. Proposed conceptual biometric shoe framework.
&
→ Conceptually (PP monitoring, model evaluation, and real-time data evaluation)
&
Conceptual only
→ Hardware (In-shoe sensors (temperature, pressure, humidity, shear), 3D foot scanning + 3D printing for custom fit) 
→ Bluetooth connectivity + mobile phone integration. 
→ Cloud infrastructure for ML analysis. 
→ ANN and supervised ML models
\\
\end{longtable}
}

Across the eight studies selected, neural networks and deep learning models were the most commonly applied, utilized in six studies overall. These included CNN–GRU, Autoencoder–RF, and LSTM–XGBoost pipelines \citep{rajagopal2025personalized, resch2025improving}, CNN+FCNN with Bayesian optimization \citep{lu2025parametric}, backpropagation neural networks (BPNN) \citep{behuranew}, a multilayer perceptron (MLP) \citep{zhang2024prediction}, and a conceptual ANN framework \citep{jones2019biometric}. Two studies applied tree-based classifiers, using Random Forest with SHAP explainability \citep{o2024ai} and CART decision trees for orthoses prescription \citep{jung2022decision}. Notably, two studies also integrated generative AI to synthesize footwear designs in a human-in-the-loop process \citep{rajagopal2025personalized, resch2025improving}. 

Most datasets were collected as part of the studies (n = 5), including data such as pressure heatmaps, orthosis images, finite element simulations, patient anthropometry, and ink footprints \citep{rajagopal2025personalized, resch2025improving, lu2025parametric, behuranew, zhang2024prediction}. One study used a large public pes planus dataset \citep{jung2022decision}, and one collected occupational plantar pressure data \citep{o2024ai}, while another study was conceptual with no empirical dataset \citep{jones2019biometric}. In one study, collected footprint–pressure pairs were augmented tenfold through synthetic embedding \citep{zhang2024prediction}. 

Decision logic was categorized into four groups. Prediction and optimization models were used in three studies to iteratively adjust footwear or insoles until plantar pressure thresholds were met \citep{lu2025parametric, behuranew, zhang2024prediction}. Classification models featured in three studies, where hierarchical models mapped gait, foot disease, or occupational tasks \citep{rajagopal2025personalized, resch2025improving, o2024ai}. One study applied rule-driven prescription using decision trees \citep{jung2022decision}, while another conceptual study proposed an ANN-driven biometric shoe for ulcer risk alerts \citep{jones2019biometric}. Notably, two studies overlapped categories, combining classification with generative AI for collaborative footwear design \citep{rajagopal2025personalized, resch2025improving}. 

Evaluation criteria were divided between model performance and footwear outcomes. Model accuracy was reported in five studies, reaching 92–98\% in hybrid CNN pipelines \citep{rajagopal2025personalized, resch2025improving}, 80–86\% for Random Forest \citep{o2024ai}, 80.16\% for CART \citep{jung2022decision}, and AUC up to 95.1\% for MLP \citep{zhang2024prediction}. Regression-based evaluations appeared in two studies, with BPNN achieving $R^2 = 0.99$ and MAE of 0.18–0.30 \citep{behuranew}, while the MLP reported MAE between 3.57–5.51\% \citep{zhang2024prediction}. Explainability was assessed through SHAP in one study \citep{o2024ai}, while one remained conceptual without empirical evaluation \citep{jones2019biometric}. Footwear evaluation was reported in five studies, with plantar pressure reduction outcomes in four \citep{lu2025parametric, behuranew, zhang2024prediction, jones2019biometric}, usability and patient feedback in two \citep{rajagopal2025personalized, resch2025improving}, and reduced design time in two \citep{lu2025parametric, behuranew}. Overlaps were common, as several studies reported both model accuracy and footwear outcomes, notably \citep{behuranew} and \citep{zhang2024prediction}, which linked predictive performance directly to clinical validation. 

Overall, the evidence shows that ML applications in offloading footwear prescription are dominated by neural network and deep learning approaches supported by proprietary datasets. Most systems focus either on predicting PP distributions or classifying gait and foot conditions, with evaluation criteria combining traditional ML metrics (accuracy, AUC, regression error) with clinically relevant measures (PP reduction, usability, and design efficiency). The frequent overlaps across categories demonstrate how ML in this field integrates both computational performance and practical footwear validation to ensure clinical relevance.

These findings provide valuable insights for the development of CDSSs in offloading footwear prescription. By demonstrating the feasibility of applying diverse ML models to predict PP distributions, classify gait or foot conditions, and optimize footwear design, these studies establish a foundation of computational tools that can be systematically embedded into CDSS architectures. Moreover, the integration of model evaluation metrics and clinically relevant outcomes, such as PP reduction and patient usability, highlights how CDSSs can be designed to combine algorithmic performance with real-world therapeutic efficacy.

\section{Roadmap for Developing a DFU Footwear CDSS}
\label{sec:roadmap}
Review of guidelines, knowledge-based systems, and ML applications highlights both the potential and the current challenges of decision-making in offloading footwear prescription. To translate these insights into a reliable CDSS, a structured roadmap is discussed in this section. This roadmap identifies five critical components: the minimum viable dataset, the decision architecture, the output specification format, validation and evaluation, and integration pathways. Together, these components provide the foundation for building scalable, transparent, and patient-centered systems. Each component is grounded in the literature and paired with its associated challenges, ensuring that the roadmap reflects both opportunities and barriers in the field. Finally, these findings form an iterative, adaptive framework to guide CDSS development in offloading footwear for DFU.

\subsection{Minimum Viable Dataset}
Across literature, PP measurement was the most common input, consistently used in guidelines and protocols for footwear validation \citep{bus2011evaluation, schaper2024practical}. Morphological and biomechanical features such as foot shape, deformities, and gait parameters were also emphasized \citep{ahmed2023ai, lopez2022clinical, zwaferink2020optimizing}. Knowledge-based systems extended this dataset with multimodal sensor streams (temperature, humidity, motion) \citep{moulaei2025usability, bencheikh2018low}, and importantly, incorporated adherence monitoring through wearable sensors and behavioral interventions \citep{van2025short, cay2025smartboot, park2023smart}. These studies highlight that adherence data, capturing whether patients consistently wear prescribed footwear, is essential for understanding real-world effectiveness and personalizing recommendations. ML studies further expanded data sources with simulated datasets \citep{lu2025parametric}, low-cost ink footprints \citep{zhang2024prediction}, and large annotated cohorts \citep{jung2022decision}.  

\textit{Challenge:} Despite these advances, the literature highlights fragmented and inconsistent datasets, often siloed across devices and study contexts \citep{australia2016australian, moulaei2025usability}. A CDSS requires a standardized minimum dataset that unifies biomechanical, morphological, clinical, and behavioral information (including adherence) to ensure interoperability, reproducibility, and clinical relevance across healthcare settings.  

\subsection{Decision Architecture}
Guidelines have largely relied on consensus-based thresholds, such as PP $\leq$200 kPa or $\geq$25\% reduction \citep{bus2011evaluation, bus2020guidelines}. Knowledge-based systems build on this foundation by applying predefined rules and continuous monitoring for adherence and biomechanical safety \citep{najafi2017smarter, ghazi2024design}. ML approaches introduce predictive classification of gait and pathology \citep{rajagopal2025personalized, o2024ai}, optimization for footwear design \citep{lu2025parametric}, and even generative AI pipelines with clinician oversight \citep{resch2025improving}.  

\textit{Challenge:} While predictive performance is strong, clinicians express limited trust in automation due to black-box ML models and lack of interpretability \citep{o2024ai}. This is particularly important as clinical users require transparency to understand why a system recommends a specific footwear feature or generates an alert. The field of explainable AI (XAI) offers tools to bridge this gap by identifying how input features influence outputs, thereby increasing user confidence and accountability. For example, a study applied the SHAP (SHapley Additive exPlanations) technique to a Random Forest model classifying occupational plantar pressure patterns, which enabled clinicians to interpret which factors (e.g., peak plantar pressure and center of pressure(CoP) variability) were driving the model’s decisions \citep{o2024ai}. A hybrid decision architecture, combining rule-based reasoning for transparency, optimization methods for iterative personalization, and ML models for adaptive predictions with embedded explainability mechanisms such as SHAP offers a balanced pathway. This approach ensures that CDSSs not only achieve high predictive accuracy but also deliver clinically interpretable and trustworthy outputs.

\subsection{Output Specification Format}
Outputs in the literature vary widely, reflecting differences in study design and system maturity. At the broadest level, guidelines provide general footwear recommendations such as wearing protective, extra depth/width shoes or custom insoles \citep{van2018diabetic, australia2016australian}, but these lack feature-level specificity needed for design and prescription. Knowledge-based systems move a step further, defining detailed design rules that translate clinical findings into modifiable footwear features. For instance, parameters such as rocker radius, heel height, apex position, and insole stiffness are systematically adjusted to achieve targeted plantar pressure reductions \citep{malki2024plantar, davia2011shoes}. Some even integrate geometric rules and CAD/3D printing workflows to ensure reproducible outputs \citep{mancuso20233d, bernabeu2013cad}.  

Machine learning applications extend this trajectory by generating full design prototypes and adaptive modifications. Generative AI systems propose multiple footwear design options, which are then curated with clinician input \citep{rajagopal2025personalized, resch2025improving}, while optimization models directly output insole configurations with predicted biomechanical performance \citep{lu2025parametric}. Conversely, other systems focus less on design parameters and instead provide real-time alerts to patients and healthcare providers (e.g., warnings for elevated plantar pressure, temperature hotspots, or low adherence) without necessarily specifying actionable footwear modifications \citep{chatwin2021intelligent, najafi2017smarter, mbue2022benefits}.  

\textit{Challenge:} This heterogeneity often leaves outputs either too broad to be useful (e.g., “wear protective footwear”) or too technical and fragmented for direct clinical adoption (e.g., raw PP maps without prescribed changes). To support clinical decision-making, CDSSs should adopt structured output specification formats, translating findings into clear, actionable feature groups. For example, “rocker sole at 95° apex angle with predicted 30\% forefoot pressure reduction.” Such formats link design parameters directly to biomechanical outcomes, enhancing interpretability for clinicians, ensuring reproducibility across systems, and making outputs both clinically relevant and technically actionable.   

\subsection{Validation and Evaluation}
The literature reveals wide variation in how decision systems for offloading footwear prescription are validated and evaluated. Guidelines primarily relied on systematic reviews and expert consensus, offering evidence-informed recommendations but with limited patient-level validation \citep{schaper2024practical, bus2020guidelines}. Protocols extended this by incorporating in-shoe PP testing and iterative adjustments, enabling biomechanical validation but largely within controlled settings \citep{giacomozzi2013learning, bus2011evaluation}. Knowledge-based systems further introduced adherence checks, usability surveys, and remote monitoring accuracy, demonstrating feasibility in real-world use \citep{cay2025smartboot, park2023smart, hemler2023intelligent}. ML studies contributed additional layers by reporting computational metrics such as model accuracy (92–98\% for CNN pipelines), regression errors ($R^2 = 0.99$, MAE 0.18–0.30 for BPNN), and explainability with SHAP \citep{o2024ai, behuranew, rajagopal2025personalized}, often combined with biomechanical outcomes like plantar pressure reductions up to 48\% \citep{lu2025parametric, zhang2024prediction}.  

Despite this diversity, validation and evaluation practices remain fragmented. Guidelines emphasize evidence synthesis, protocols focus on biomechanics, knowledge-based systems assess usability and adherence, and ML studies highlight predictive accuracy. However, few approaches embed both continuous patient-level safety loops and comprehensive system-level evaluation. As a result, most prototypes lack mechanisms for adaptive refinement or demonstration of long-term clinical benefit, such as ulcer recurrence or cost-effectiveness \citep{najafi2018cost, mancuso20233d}.  

\textit{Challenge:} The current landscape is siloed, with studies emphasizing narrow aspects of performance. What is missing is an integrated framework that combines immediate patient-level checks with broader performance assessment. A robust CDSS should therefore adopt a multi-tiered validation and evaluation framework. At the patient level, a continuous safety loop is needed, linking prescription outputs to PP re-testing, adherence monitoring, usability feedback, and adaptive refinement—with an emphasis on actively involving patients in the co-design process. At the system level, evaluation must align biomechanical measures (e.g., PP reduction, gait dynamics), process outcomes (e.g., decision time, adherence accuracy, usability), and clinical endpoints (e.g., ulcer prevention, recurrence, mobility, cost-effectiveness). Such integration would ensure both short-term safety and long-term credibility, providing the evidence base required for regulatory approval and real-world adoption.

\subsection{Integration Pathways}
Integration into clinical practice is a recurring theme across the literature, though approaches vary in maturity. Guidelines emphasize the importance of consistency across care settings, ensuring that recommendations are applicable from specialist hospital clinics to community podiatry and rehabilitation services \citep{van2018diabetic, australia2016australian}. This reflects a recognition that offloading footwear prescription must be standardized to support equitable care and benchmarking across regions.  

Knowledge-based systems illustrate more direct examples of integration by embedding decision logic into digital health infrastructures. These include cloud-based dashboards that allow healthcare professionals to remotely monitor adherence and plantar pressure data, smartwatch-based alert systems that engage patients directly, and telemonitoring platforms that extend care beyond the clinic \citep{ghazi2024design, mbue2022benefits, cay2025smartboot}. Such implementations demonstrate the feasibility of integrating CDSS outputs into day-to-day patient management and highlight the value of linking clinicians and patients in real time.  

ML studies show additional pathways by demonstrating adaptability across both high-resource and low-resource contexts. For example, advanced gait laboratory infrastructure with motion capture and FE simulation supports high-fidelity model development \citep{lu2025parametric}, while low-cost data collection methods such as ink footprint analysis expand feasibility in community or resource-constrained environments \citep{zhang2024prediction}. Together, these studies suggest that ML-enabled systems could be scalable across settings if designed with modularity in mind.  

\textit{Challenge:} Despite these examples, many systems remain siloed prototypes with limited interoperability and poor workflow alignment, creating additional burden rather than clinical support \citep{ghazi2024design, park2023smart}. Integration into CDSS development requires prioritizing seamless embedding within existing care pathways—podiatry workflows, rehabilitation services, and telehealth—while ensuring minimal disruption to clinical routines. Interoperability with electronic health records and scalability across diverse healthcare contexts will be critical for real-world adoption.  

\subsection{Proposed Framework for CDSS Development}
\begin{figure}[H]
\centering
\includegraphics[width=\linewidth]{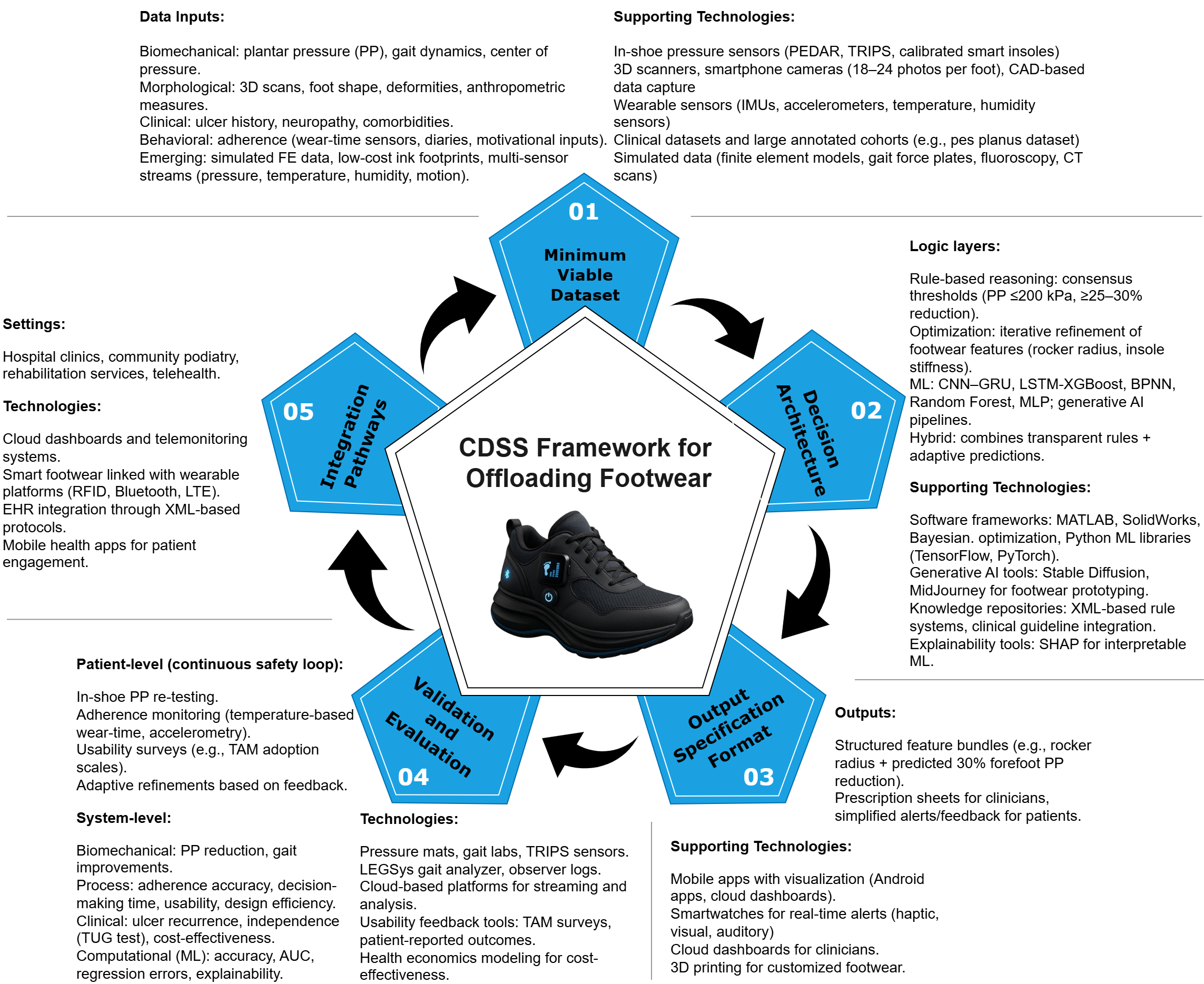}
\caption{Framework for developing a CDSS for offloading footwear prescription for DFU patients}
\label{fig:framework}
\end{figure}

Drawing of components identified in the roadmap, we propose an overall framework that logically integrates these into an iterative development cycle for CDSSs in offloading footwear prescription (Figure~\ref{fig:framework}). The framework starts with a standardized minimum viable dataset, combining biomechanical, morphological, clinical, and adherence data as the foundation for decision-making. A layered decision architecture then consolidates rule-based reasoning, optimization methods, and ML to generate outputs. the next layer expresses these outputs in structured feature groups, translating system recommendations into clinically actionable footwear specifications (such as, rocker angle, insole stiffness, predicted PP reduction).  

Furthermore, each prescription is linked to a unified validation and evaluation framework. At the patient level, this includes continuous safety loops through in-shoe PP re-testing, adherence monitoring, usability feedback, and adaptive refinement. At the system level, it integrates biomechanical, process, and clinical outcomes into a multi-tiered evaluation structure, ensuring regulatory credibility and real-world effectiveness. Validated outputs are then embedded into integration pathways covering hospital clinics, community podiatry, and telehealth services, ensuring scalability and workflow compatibility.  

This framework further emphasizes that this CDSS development is not a linear process but a cyclical, adaptive system, where datasets, decision logic, outputs, and evaluations iteratively inform one another. By combining computational innovation with continuous and multi-level evaluation, the framework ensures that future CDSSs for offloading footwear prescription remain transparent, safe, patient-centered, and scalable.  

\section{Conclusion}
\label{sec:conclusion}

This review highlights the diverse decision-making approaches currently applied to offloading footwear prescription, spanning twelve guidelines and protocols, 25 knowledge-based systems, and eight machine learning applications. Across these studies, PP thresholds and adherence emerged as the most consistent targets, with rule-based and sensor-driven logic dominating knowledge-based systems, and predictive or generative models characterizing recent works on ML. However, footwear and system evaluation practices remain fragmented: with guidelines emphasizing evidence synthesis, protocols focus on biomechanics, knowledge-based systems test usability and adherence, while ML studies largely report model accuracy without long-term clinical validation. From this synthesis, we propose a structured roadmap and CDSS framework that integrates five critical components: a standardized minimum dataset, hybrid decision architecture, structured feature-level outputs, continuous validation and evaluation, and integration pathways into clinical workflows. Together, these elements provide a pathway for translating existing prototypes into scalable, explainable, and patient-centered CDSSs. Future research should prioritize developing interoperable datasets, embedding explainability into ML models, and aligning evaluation with clinical outcomes such as ulcer recurrence, adherence, and usability. Establishing standardized methodologies for validation and evaluation will not only facilitate comparability across studies but also accelerate regulatory readiness and clinical adoption. Ultimately, advancing CDSSs in this domain holds the potential to reduce clinician burden, improve adherence, and deliver safer, more personalized footwear prescriptions for DFU patients.  

\backmatter


\section*{Declarations}

\subsection*{Ethics approval and consent to participate}
Not applicable.

\subsection*{Consent for publication}
Not applicable.

\subsection*{Availability of data and materials}
Not applicable.

\subsection*{Competing interests}
The authors declare that they have no competing interests. Ashad Kabir serves as a member of the editorial board of BMC Medical Informatics and Decision Making.

\subsection*{Funding}
This research was supported by the Australian National Industry PhD Program (Award Reference No. 34954). The funding body had no role in the study design, data collection, analysis, interpretation, or manuscript writing.

\subsection*{Authors' contributions}
\textbf{Kunal Kumar} conceptualized the study, conducted the review, and led the manuscript writing.  
\textbf{Muhammad Ashad Kabir} contributed to the study design, conceptualization, methodology, and visualization. He also reviewed and edited the manuscript, supervised and administered the project, and secured funding.  
\textbf{Luke Donnan} reviewed and edited the manuscript, providing critical feedback on clinical implications.  
\textbf{Sayed Ahmed} contributed to the discussion on industry applications and provided insights into real-world implementation.  
All authors read and approved the final manuscript.
\subsection*{Acknowledgments}
Not Applicable

\bibliography{bibliography}

\end{document}